\documentclass{article}
\usepackage{ijcai22}

\usepackage{times}
\usepackage{multirow}

\usepackage{soul}
\usepackage{url}
\usepackage[hidelinks]{hyperref}
\usepackage[utf8]{inputenc}
\usepackage[small]{caption}
\usepackage{graphicx}
\usepackage{amsmath}
\usepackage{booktabs}
\usepackage{amsfonts}
\usepackage{amsmath}

\DeclareMathOperator*{\argmin}{argmin}
\DeclareMathOperator*{\sigmoid}{sigmoid}

\newcommand{\comment}[1]{}
\usepackage{arydshln}

\usepackage{graphicx}
\usepackage{subcaption} %to have subfigures available
\newsavebox{\measurebox}

\title{RLSEP: Learning Label Ranks for Multi-label Classification}

\author{
Emine Dari\footnote{Equal Contribution}$^1$\and
V. Bugra Yesilkaynak$^*$\footnote{Contact Author}$^2$\and
Alican Mertan$^{3}$\And
Gozde Unal$^1$\\
\affiliations
$^1$Istanbul Technical University, Turkey\\
$^2$Technical University of Munich, Germany\\
$^3$University of Vermont, USA\\
\emails
\{dari18, gozde.unal\}@itu.edu.tr,
bugra.yesilkaynak@tum.de,
alican.mertan@uvm.edu 
}

\begin{document}

\maketitle

\begin{abstract}
Multi-label ranking maps instances to a ranked set of predicted labels from multiple possible classes. The ranking approach for multi-label learning problems received attention for its success in multi-label classification, with one of the well-known approaches being pairwise label ranking. However, most existing methods assume that only partial information about the preference relation is known, which is inferred from the partition of labels into a positive and negative set, then treat labels with equal importance. In this paper, we focus on the unique challenge of ranking when the order of the true label set is provided. We propose a novel dedicated loss function to optimize models by incorporating penalties for incorrectly ranked pairs, and make use of the ranking information present in the input. Our method achieves the best reported performance measures on both synthetic and real world ranked datasets and shows improvements on overall ranking of labels. Our experimental results demonstrate that our approach is generalizable to a variety of multi-label classification and ranking tasks, while revealing a calibration towards a certain ranking ordering.
\end{abstract}

\section{Introduction}
Multi-label ranking is a complex prediction task realized by the two important learning problems, multi-label classification and label ranking. As a summary of the detailed comparison given by \cite{Zhou2014ATO}, the aim of multi-label classification is to separate a set of labels into two by their relevance to the instance. Unlike multi-class classification, where an instance is associated with a single label only, multi-label classification associates instances with a subset of labels. The second problem, label ranking, concerns with learning a label preference over a set of labels and order all of them. The different objectives of these supervised learning problems construct the sub-problems of the multi-label ranking algorithms, where the target is to predict the relevant subset of all possible labels and represent their order of relevance to the instance by providing a ranking among them.

\begin{figure}[t]
    \centering
    \begin{tabular}{c}
    \includegraphics[width=60mm]{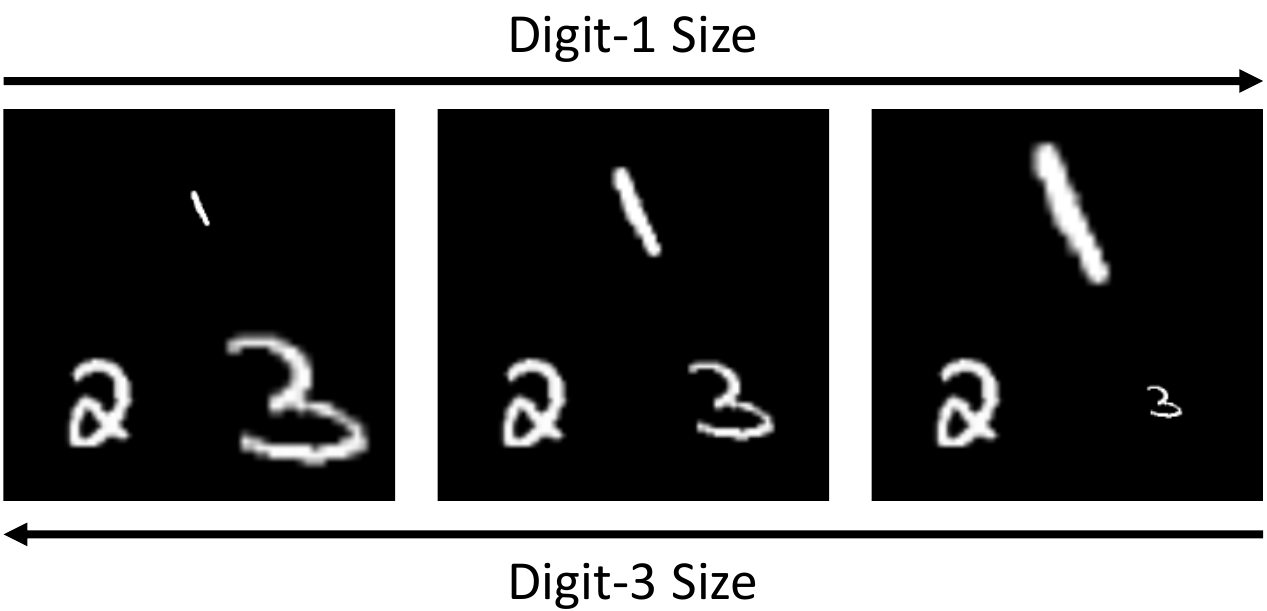} \\
    \hspace*{-0.25in}
    \includegraphics[width=65mm]{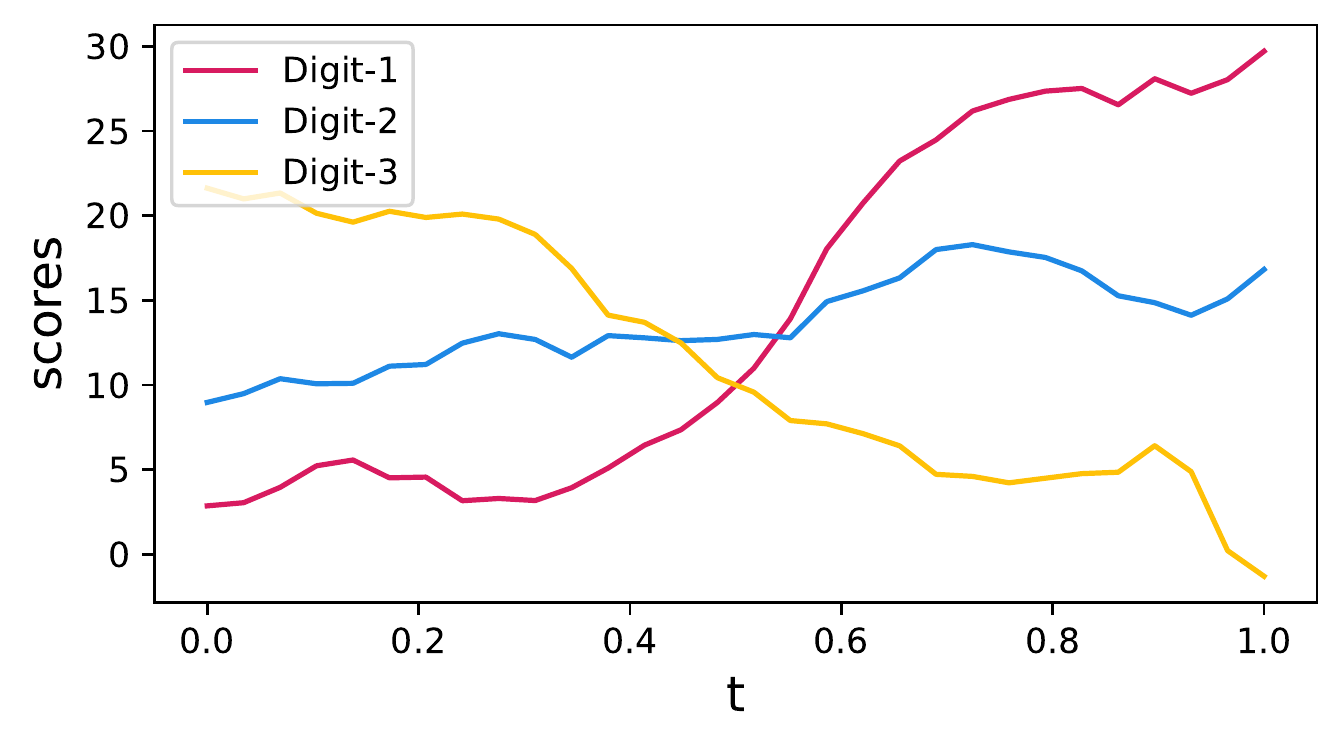}
    \end{tabular}
    \caption{Illustration of changes in the predicted scores for each digit as their sizes change. The predicted scores are direct outputs of a network trained with RLSEP. While training the network, the sizes of the digits are given as their importance for the image. RLSEP successfully produces consistent scores for the three digits, as it can be observed in the plot. The figure consists of 30 frames, as t increases, "Digit-1" gets bigger and "Digit-3" gets smaller.}
    \label{fig:motivational}
\end{figure}

On a general note, the difference between learning to rank from label preferences and object preferences should be clearly distinguished. In object ranking \cite{Cohen1997LearningTO}, the goal is to learn a ranking function that takes a subset of a predefined class of objects as the input, and produces a permutation of objects which provides a relation of preferences, conducted from the indices that objects are placed in. Object ranking has applications including information retrieval \cite{BaezaYates} or collaborative filtering \cite{Harrington2003}, where the goal is to serve the information in an order representing objects’ relevance to the query made. In label ranking, a preference relation over a finite set of labels is produced for each instance, where the order in ranking is conducted from the position in the output permutation \cite{hullermeier_2008}. This paper concentrates on learning a ranking of predicted labels that implies their relevance to an instance and investigates the interclass relations.

For multi-label predictions, even when the prediction is a non-ranked subset of all labels, the cruciality of accurate ranking stems from its impact on the final prediction set when there is no fixed constraint on how many labels can be associated with the sample. For this reason, there is an increasing number of approaches to multi-label ranking, as well as its applications to a variety of different disciplines. A study on multi-label classification by \cite{Ioannou2010ObtainingBF}, obtained ranking information from predicted scores of each class to learn a dynamic threshold to output a better bipartition with respect to the labels’ relevance. In object recognition, \cite{Bucak2009EfficientMR} utilized multi-label ranking to order the relevance of objects present in images.

According to our observations, although there are many studies on the multi-label ranking problem, existing approaches train models with labels of equal importance and label ranker functions are optimized with partial information that is deduced from the comparison of positive and negative labels only. In addition, prediction of the relevant label set is performed by fixed label counts or thresholds. They lack bringing a solution to one of the challenges of multi-label learning, which is label importance \cite{Dery2021MultilabelRM}. Label importance means that the associated labels are not equally important when inferring to the instance, and for this reason, correctly ordering the associated labels matters. 

In this work, we address the aforementioned problems and present our contributions by experiments in the context of ranked multi-label image classification. We propose a novel loss function that is tailored for optimizing rankings according to ground truth label orders. Our proposed method shows improvements on overall label ranking and obtains calibrated results with respect to conditions that determine the label orders. Notably, our method achieves the best scores on both synthetic and real world datasets with ranked labels, when evaluated with different metrics to measure both prediction accuracy and label orders.

\section{Related Work}

Ranking is studied in information retrieval literature, where there are documents to be sorted based on a  given query. The works in this literature roughly fall into three categories based on the number of items considered in the loss function. While \cite{caruana_using_1996} regresses to a target value per item, \cite{joachims_optimizing_2002,burges_learning_2005,tsai_frank:_2007} consider a pair of items at a time in their loss functions and predict a score that correctly ranks them. Lastly, \cite{xia_listwise_2008,lan_position-aware_2014} use listwise loss functions defined over permutations of items. Recently, we see ranking in computer vision problems such as relative depth estimation \cite{zoran_learning_2015,chen_learning_2019,mertan2020relative,mertan_new_2020}, reflectance estimation % narihira_learning_2015, zhou_learning_2015
\cite{zoran_learning_2015}, representation learning \cite{liu_exploiting_2019}, or image retrieval \cite{cakir_deep_2019,revaud_learning_2019}. % chen_single-image_2016,xian_monocular_2018

Learning preferences over labels instead of instances or objects was first introduced in the constraint classification framework proposed by \cite{HarPeled2002ConstraintCF}. Constraint classification investigates the relationships between labels by associating each instance with constraints that specify the order of relevance between the labels. The constraints mentioned are formed regarding the label preferences where the relation $\lambda_i > \lambda_j$, denoting that label $\lambda_i$ is more relevant to the instance than label $\lambda_j$, constructs the constraints $f_i(x) - f_j(x)>0$ or $ f_j(x) - f_i(x)<0$. These two positive and negative constraints are then expressed as training samples for a binary classifier in a high-dimensional space to obtain a weight vector consistent with the constraints. As the constraint classification algorithm creates twice as many training samples than other pairwise comparison algorithms, it is considered to be less efficient in practice \cite{Zhou2014ATO}.

Another well-known algorithm called ranking by pairwise comparison \cite{hullermeier_2008} takes each pair of labels as instances and reduces ranking to a binary problem where a model is trained for each label pair. The model learns a mapping to produce outputs in the interval [0,1], with 1 demonstrating the pair is in correct order, and 0 as the pair is in reversed order. By transforming the output interval to \{0,1\}, relations can be represented as probabilities. \cite{10.5555/1567016.1567123} extended this pairwise comparison algorithm which was initially proposed for label ranking \cite{hullermeier_2008} into a unified model using a calibrated ranking method, where a learned natural zero-point determines the relevancy cutoff, and approximates the pairwise approach to multi-label learning.

Most similar to our work, \cite{DBLP:journals/corr/LiSL17} proposed the log-sum-exp-pairwise (LSEP) loss function to rank the label pairs with a function that is smooth and easier to optimize when compared with other traditional approaches using a hinge loss \cite{Gong2014DeepCR,Weston2011WSABIESU} to train deep convolutional neural networks. %\cite{Krizhevsky2012ImageNetCW}
LSEP takes label pairs constructed by choosing one label from the positive (relevant), and one from the negative (irrelevant) set according to ground truth labels. Then LSEP enforces the model to produce a vector with positive labels in higher ranks and negative labels in lower ranks, by comparing the predicted scores for each label in different sets. However, an exactly ranked comparison was not performed, which is one of our contributions in this work.

\section{Preliminaries}

\subsection{Notations}
We formally define datasets with unranked labels as $\mathcal{D}_{ur} = \{(x_i, Y_i)\}_{i=1}^N$ where $x_i \in \mathbb{R}^d$ is the i-th input, $Y_i$ is the corresponding unranked label set. The number of labels $|Y_i| = k_i$ can be different for each input, and $Y_i \subset \mathcal{Y}$ where $\mathcal{Y} = \{y_i\}_{i=1}^K$ is the set of all possible labels. Whereas we define datasets with ranked labels as $\mathcal{D}_{r} = \{(x_i, R_i)\}_{i=1}^N$ where $R_i = \{(y_j, r^{(i)}_j)\}_{j=1}^K$ and $r^{(i)}_j \in \mathbb{N}$ is the rank of $y_j$ for the i-th input. That is to say, a label set in a ranked dataset has all possible labels with an associated rank.

Let us assume for each $x_i$ in the input space of $D_r$, the exact importance of each $y_j \in \mathcal{Y}$ exists as a numerical value denoted as $s_j$, and it only depends on the input instance $x_i$. We do not have direct access to the value of $s_j$, but we know if $s_u > s_v$ or $s_u \sim s_v$ for $u, v \in \{1, .., K\}$. We generate ranks $r_j \in \mathbb{N}$ from these importance relationships such that if $s_u > s_v$ then $r_u > r_v$, and if $s_u \sim s_v$ then $r_u = r_v$. For each ranked label set $R_i$ in $D_r$ the least important labels' ranks are assigned to 0 (usually the negative labels), then the remaining label ranks are iteratively assigned by increasing the ongoing minimum by one. One can also consider an unranked dataset as a ranked dataset, where $r_j = 1$ if $y_j \in Y_i$, and $r_j = 0$ otherwise. It should be noted that the difference between ranks is only meaningful for ordering and does not yield the difference of importance. For example, let $S = \{(y_1, s_1), (y_2, s_2), (y_3, s_3)\}$ be the label set with importance values for an arbitrary input. Even if $s_2 >> s_1 > s_3$ holds, the corresponding ranks have the relation $r_2 > r_1 > r_3$ and $r_2 - r_1 = r_1 - r_3 = 1$ since we do not know the exact importance values.

% $f(x)$ is a function $f: \mathbb{R}^d \rightarrow \mathbb{R}^K$ which takes an input from the input space and outputs the corresponding label logits. If used in the form $f(x; \theta)$, $\theta$ denotes the learnable parameters of $f$. $\sigma(z)$ denotes the softmax function.
% \begin{equation}
%     \sigma(z)_i = \dfrac{e^{z_i}}{\sum_{j=1}^K e^{z_j}}, z \in \mathbb{R}^K
% \end{equation}
% $\sigma(f(x_i; \theta)) = \hat{y}_i$ yields the predicted class probabilities for $x_i$.

\subsection{Label Rank Prediction with Cross Entropy}

A simple approach to the label ranking problem is to use cross entropy. For an unranked dataset $D_{ur}$, we can denote $p(y_j|x_i) = 1/|Y_i|$ if $y_j \in Y_i$, and $p(y_j|x_i) = 0$ otherwise, with $i \in \{1, .., N\}, j \in \{1, .., K\}$. Then we can maximize the cross entropy between the ground truth and predicted probabilities with respect to $\theta$: 
\begin{equation}
    \max_\theta \sum_{i=1}^N \sum_{j=1}^K p(y_j | x_i) \log \sigma ( f(x_i; \theta) )_j,
\end{equation}
where $\sigma( f(x_i; \theta) )_j$ denotes the predicted probability of label $j$ for data instance $i$, and $\sigma$ denotes softmax function. In our work, we call this Cross Entropy (CE) approach, which however cannot be used directly on ranked datasets. In such cases though, we convert the ranked dataset into an unranked one by changing every $r_j > 0$ to $r_j = 1$, and if $r_j = 1$ then $y_j \in Y_i$.

\begin{figure*}[ht]
\centering
\resizebox{0.8\textwidth}{!}{
\includegraphics[width=\linewidth]{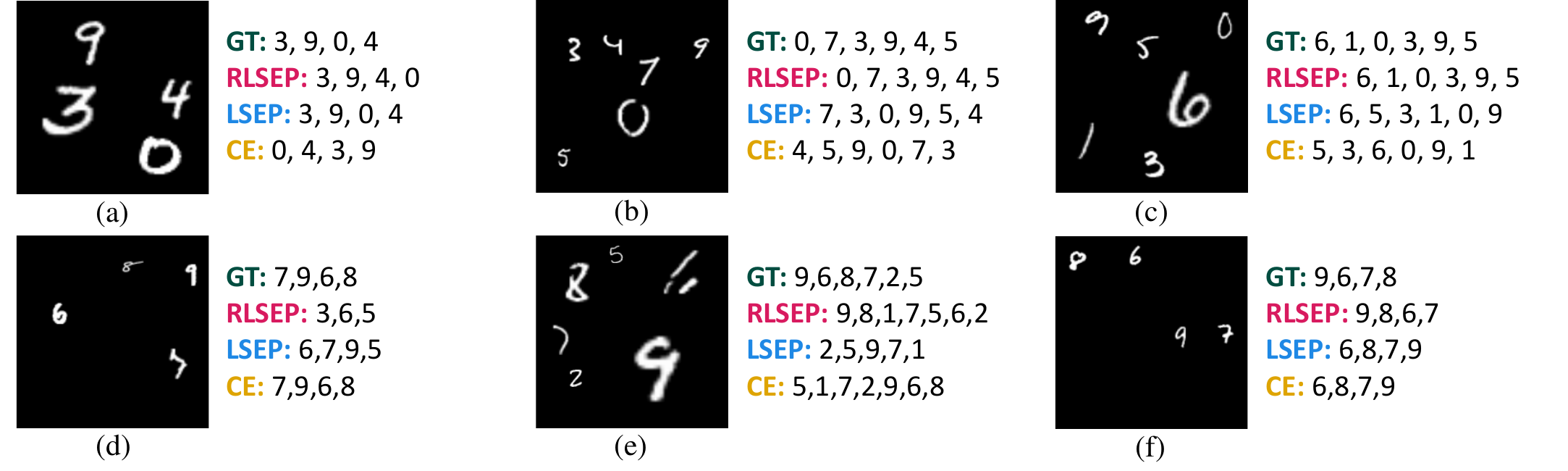}
}
\caption{Illustrative Multi-label Ranking results on Ranked MNIST. GT denotes the ground truth ranking. The top row (a-c) shows some successful results of RLSEP where exact ranking is predicted, whereas (d-f) depicts failures in challenging cases. RLSEP, LSEP and CE denote different scoring functions. To get a full multi-label ranking output, a threshold network is used after the scoring for all scoring functions (Sec~\ref{subsec:label_selection}). Label importance is given in descending order from left to right for the ground truth and predictions.}
\label{fig:mnist_results}
\end{figure*}

\subsection{Label Rank Prediction with LSEP}

Another alternative to the label ranking problem is the \textit{Log-Sum-Exp Pairwise (LSEP)} loss \cite{DBLP:journals/corr/LiSL17}. For an unranked dataset, LSEP tries to learn a score function such that for scores $s = f(x_i; \theta)$, $s_u > s_v$ if $y_u \in Y_i$ and $y_v \notin Y_i$ for any $(u, v) \in \{1, .., K\}^2$, meaning a positive label should have a higher score compared to that of a negative one. A simple way of formulating the problem is:
\begin{equation}
\label{equ:nonconvexlsep}
\min_\theta \sum_i \sum_{y_u \in Y_i} \sum_{y_v \notin Y_i} \mathbb{I}[ f(x_i; \theta)_u < f(x_i; \theta)_v ].
\end{equation}
LSEP is a smooth function utilizing the idea in Eq.~(\ref{equ:nonconvexlsep}), which is formulated as:
\begin{equation}
l_{lsep} = log\left( 1 + \sum_{\phi(Y_i; t)} exp(f(x_i; \theta)_v - f(x_i; \theta)_u) \right)
\end{equation}
where $$\phi(Y_i; t) \subseteq \{(u, v) | y_u \in Y_i \wedge y_v \notin Y_i, (u,v) \in \{1, .., K\}^2\}$$
and $|\phi(Y_i; t)| = t$, where t denotes the size of the random subset. Typically, the pairwise comparison has $\mathcal{O}(K^2)$ time and memory complexity, however using a subset of the pairs makes the function scale linearly with an increasing number of classes.

\begin{figure}[ht]
\centering
\hspace*{-0.12in}
\begin{tabular}{c}
\includegraphics[width=0.6\linewidth]{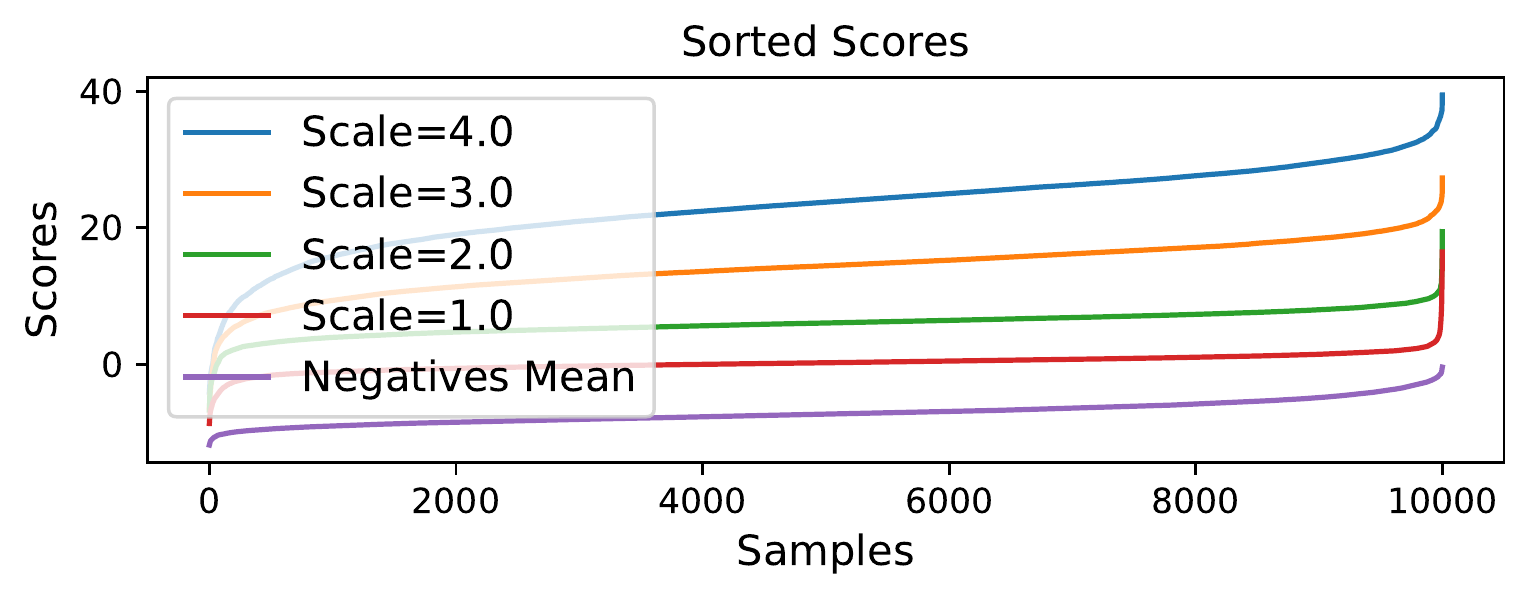} \\
\includegraphics[width=0.6\linewidth]{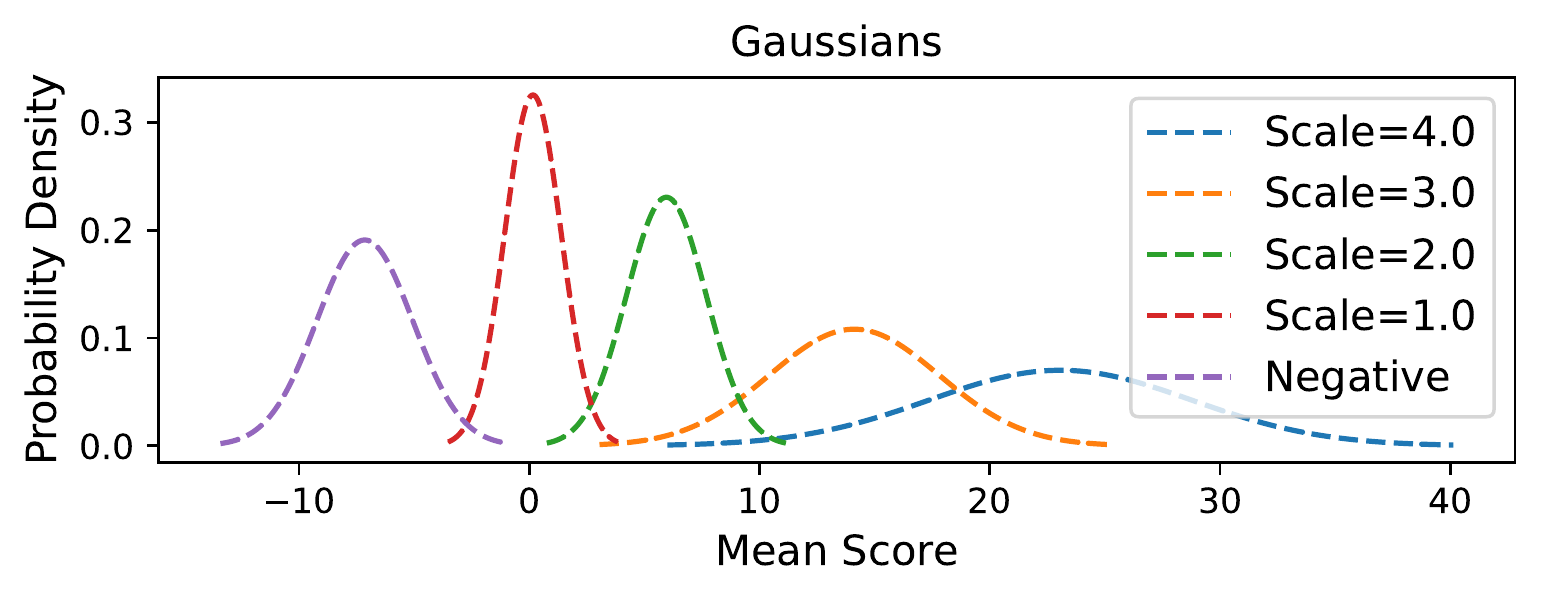} \\
\hspace*{0.12in}
\includegraphics[width=0.5\linewidth]{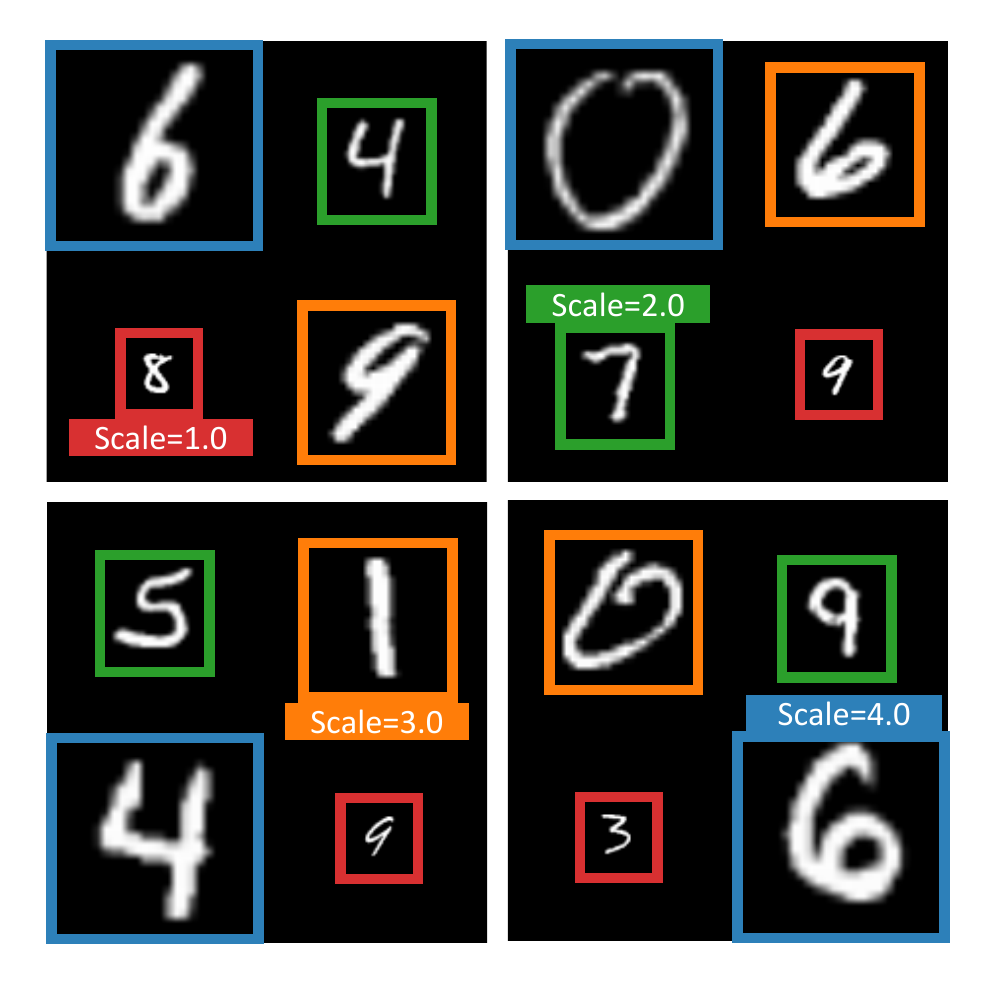}
\end{tabular}
\caption{Results of the calibration test on scores predicted by RLSEP. The uppermost plot shows scores for 10,000 images, each image has four distinct scaled digits as can be seen in the bottom figure, for visual convenience they are sorted in the x-axis. "Negatives mean" is the mean of scores of digits which are not present in an image. The middle plot shows the same scores fitted to Gaussian distributions.}
\label{fig:calibration}
\end{figure}

\subsection{Label Selection}
\label{subsec:label_selection}
With Cross Entropy and LSEP, it is possible to learn multi-label scores, but these scores do not naturally convert into positive and negative labels. We use the dynamic multi-threshold method \cite{DBLP:journals/corr/LiSL17} to address this problem. Let $f_{feat}(x;\theta)$ be the feature extractor, meaning the last layer before the classification head, where $\theta$ is optimized using any scoring loss (e.g. CE, LSEP). We train a small threshold network $g(; \gamma): \mathbb{R}^q \rightarrow \mathbb{R}^K$ via
\begin{equation}
\gamma^* = \argmin_\gamma \sum_{i=1}^N \sum_{j=1}^K Y_{i,j} \log(\delta_{i,j}) + (1 - Y_{i,j}) \log(1 - \delta_{i,j})
\end{equation}
where $\delta_{i,j} = \sigmoid(f(x_i; \theta)_j - g(f_{feat}(x_i; \theta); \gamma)_j)$ and $Y_{i,j} \in \{0, 1\}$. That is to say, for any input, the additional network takes features from the pre-trained network and predicts thresholds to decide if a label is positive or negative given its score.

\section{Multi-label Ranking with RLSEP}

Let $\mathcal{D}_{r} = \{(x_i, R_i)\}_{i=1}^N$ be our dataset, where $R_i = \{(y_j, r^{(i)}_j)\}_{j=1}^K$ and $r^{(i)}_j \in \mathbb{N}$ is the rank of $y_j$ for the i-th input. Each label has an associated rank value, these rank values are used only to determine a pairwise magnitude relation, and their difference or ratio does not indicate any further information.

Our goal is to learn a label ranking function $f(x)$ for multi-label ranking. The function $f(x): \mathbb{R}^d \rightarrow \mathbb{R}^K$ produces a label-score vector. The elements of this vector can be interpreted as pseudo-significance values, that is to say the magnitude relation between each pair of label-scores are also the predicted relation between the corresponding labels, which yields the predicted ranking of the full label set $\mathcal{Y}$ for an input $x$.

\textbf{Label ranking.} We learn the score function $f(x; \theta): \mathbb{R}^d \rightarrow \mathbb{R}^K$ by solving the optimization problem:
\begin{equation}
\theta^* = \argmin_\theta \sum_i^N l(f(x_i; \theta), R_i) + \mathcal{R}(\theta)
\end{equation}
where $\theta$ is the parameter vector of the function $f$, and $\mathcal{R}$ is the regularization function, e.g.  the squared $L_2$ norm of $\theta$.

To learn a sorting of the labels $\mathcal{Y}$ for a given $x$, one way is to predict the pairwise magnitude relations, meaning for a ranked dataset with $x$ as the input, $R$ as the corresponding ranked label set and $r_j, \forall j$ as label ranks, we require a scoring function $f$ to produce scores which are proportional to the ranks such that: $f(x)_u > f(x)_v$ if $r_u > r_v$ and $f(x)_u < f(x)_v$ if $r_u < r_v$, hence we introduce the \textit{Ranked Log-Sum-Exp Pairwise (RLSEP)} loss:
\begin{equation}
\begin{split}
l_{rlsep} = log\left( 1 + \sum_{\zeta(R; t)} exp(f(x)_v - f(x)_u) \right)
\\
\zeta(R; t) \subseteq \{(u, v) | r_u > r_v, (u,v) \in \{1, .., K\}^2\}
\end{split}
\end{equation}
The parameter $t$ is used to decide the number of elements in the pair subset, meaning $|\zeta(R;t)| = t$ and the subset is chosen randomly as in \textit{Negative Sampling Method} introduced in \cite{DBLP:journals/corr/MikolovSCCD13}, similarly to LSEP. It should be noted that when positive labels have rank 1 and negative labels have rank 0, RLSEP becomes equivalent to LSEP. Hence, RLSEP generalizes the idea by using additional label ranks.

\textbf{Label selection.} After training the network for label ranking with RLSEP, we apply the threshold method explained in Section \ref{subsec:label_selection} to produce multi-label ranking results.

\begin{figure}[ht]
    \centering
    \includegraphics[width=0.48\textwidth]{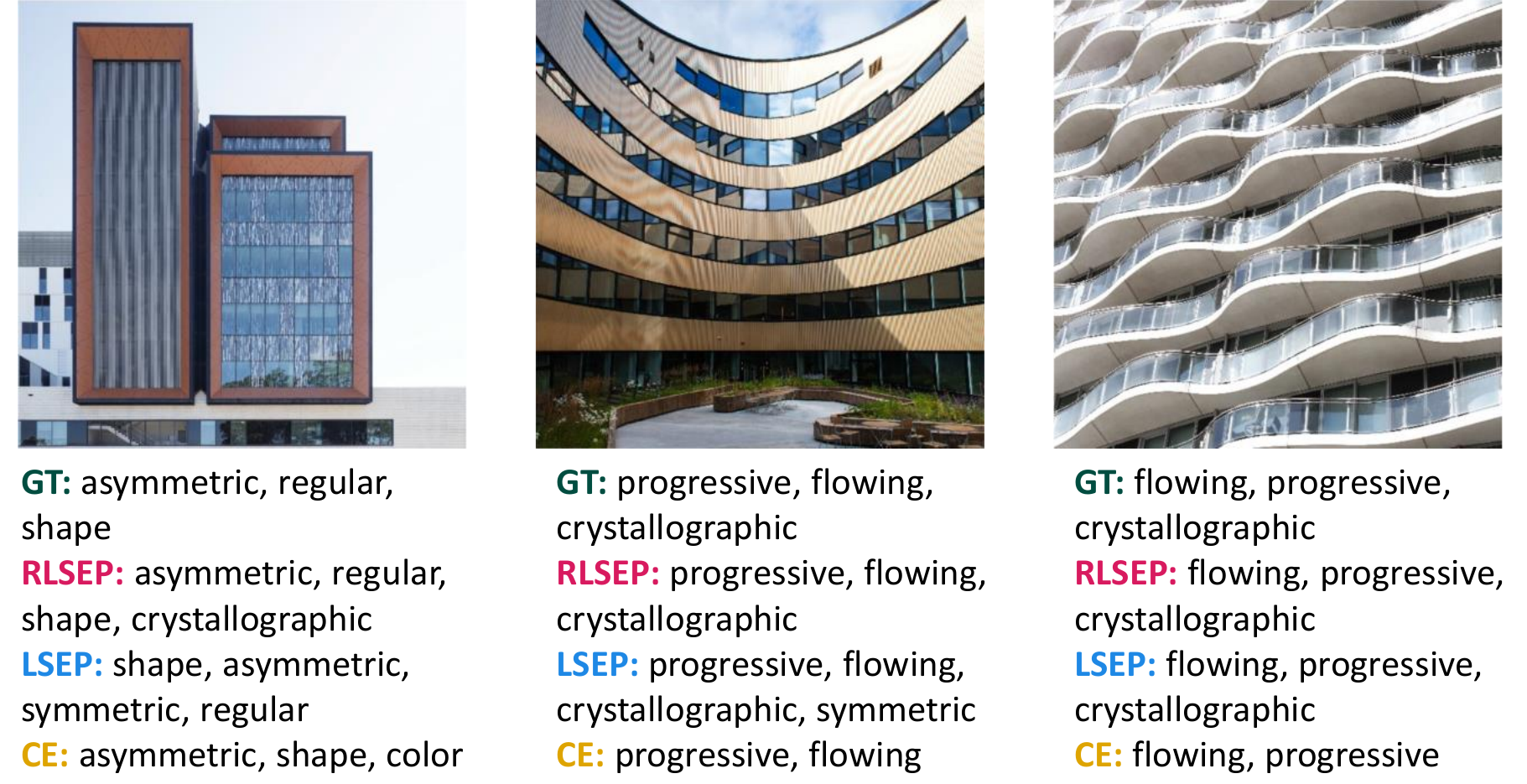} 
    \caption{Sample results from the ARC (architecture) dataset.}
    \label{fig:arc_results}
\end{figure}

\section{Experiments}

\subsection{Datasets}
We used two datasets to compare RLSEP with baseline methods and demonstrate the advantages over, as well as provide qualitative results and analyses on different experimental setups to prove the relevancy of our method.

\textbf{Architecture Dataset} The set of architectural facade images (ARC) introduced in \cite{DEMIR2021103826} was obtained upon request to be used in our experiments. ARC dataset contains 4145 annotated images, with  9 classes that represent visual design principles defined as color, isolation, shape, symmetric, asymmetric, crystallographic, regular, progressive, and flowing. Each image is associated with a maximum of 3 labels, and the order of true labels is present. Sample images are shown in Figure \ref{fig:arc_results}.

\textbf{Ranked MNIST} We created a ranked multi-class dataset from the MNIST dataset \cite{deng2012mnist} by generating images of 224x224 that contain different digits, where the number of digits in a single image vary from 3 up to 6. Each digit is scaled by a scale factor $s \sim U(1,4)$ using bilinear interpolation before being placed on the image, making their spatial dimensions vary between 28x28 and 112x112. A total of 60,000 training, 10,000 validation, 10,000 test images were generated, with the scale factors of digits determining the label order. Samples can be seen in Figure \ref{fig:mnist_results}.

% \begin{figure}[ht]
%     \begin{subfigure}[t]{0.48\textwidth}
%         \centering
%         \raisebox{-\height}{\includegraphics[width=0.24\textwidth]{arc/asym.PNG}}
%         \raisebox{-\height}{\includegraphics[width=0.24\textwidth]{arc/color.PNG}}
%         \raisebox{-\height}{\includegraphics[width=0.24\textwidth]{arc/crystal.PNG}}
%         \raisebox{-\height}{\includegraphics[width=0.24\textwidth]{arc/flow.PNG}}
%         %\raisebox{-\height}{\includegraphics[width=0.15\textwidth]{arc/isolation.PNG}}
%         %\raisebox{-\height}{\includegraphics[width=0.15\textwidth]{arc/progres.PNG}}
%         %\caption{ARCH Dataset}
%     \end{subfigure} \\
%     % second row
%     % \begin{subfigure}[t]{0.48\textwidth}
%     %     \centering
%     %     \raisebox{-\height}{\includegraphics[width=0.24\textwidth]{ranked_mnist/11.png}}
%     %     \raisebox{-\height}{\includegraphics[width=0.24\textwidth]{ranked_mnist/14.png}}
%     %     \raisebox{-\height}{\includegraphics[width=0.24\textwidth]{ranked_mnist/15.png}}
%     %     \raisebox{-\height}{\includegraphics[width=0.24\textwidth]{ranked_mnist/16.png}}
%     %     %\raisebox{-\height}{\includegraphics[width=0.15\textwidth]{ranked_mnist/17.png}}
%     %     %\raisebox{-\height}{\includegraphics[width=0.15\textwidth]{ranked_mnist/21.png}}
%     %     \caption{Ranked MNIST Dataset}
%     % \end{subfigure}
%     \caption{Sample images from the two datasets used in experiments}
%     \label{fig:arc_results}
% \end{figure}

\begin{table}[ht]
 \renewcommand{\arraystretch}{1.2}
\centering
\resizebox{0.47\textwidth}{!}{
\begin{tabular}{c|cccccc}
\multirow{2}{*}{Method} & \multicolumn{5}{c}{Ranked MNIST (Subset)}                                                 \\ \cline{2-7} 
                        & Precision & \multicolumn{1}{c|}{Recall} & F1   & \multicolumn{1}{c|}{Acc}  & \multicolumn{1}{c|}{0-1} & mAP \\ \hline
CE (R)                   & $85.5 \pm 0.7$      & \multicolumn{1}{c|}{$85.9 \pm 1.2$}   & $85.7 \pm 0.6$ & \multicolumn{1}{c|}{$87.1 \pm 0.4$} & \multicolumn{1}{c|}{$5.9 \pm 2.2$} & 79.5 $\pm$ 0.8 \\
LSEP (R)                 & $85.7 \pm 0.9$      & \multicolumn{1}{c|}{$86.1 \pm 1.0$}   & $85.9 \pm 0.6$ & \multicolumn{1}{c|}{$87.4 \pm 0.5$} & \multicolumn{1}{c|}{$5.5 \pm 1.7$} & 79.4 $\pm$ 1.2 \\
RLSEP (R)                & \textbf{95.9 $\pm$ 0.5}      & \multicolumn{1}{c|}{\textbf{97.1 $\pm$ 0.7}}   & \textbf{96.5 $\pm$ 0.5} & \multicolumn{1}{c|}{\textbf{97.0 $\pm$ 0.4}} & \multicolumn{1}{c|}{\textbf{45.4 $\pm$ 5.6}} & \textbf{93.6 $\pm$ 0.6} \\ \hline
%CE (U/S)                   & $96.2 \pm 0.7$      & \multicolumn{1}{c|}{\textbf{98.1 $\pm$ 0.4}}   & $97.1 \pm 0.5$ & \multicolumn{1}{c|}{$96.9 \pm 0.5$} & \multicolumn{1}{c|}{$77.7 \pm 4.2$} & - \\
%LSEP (U/S)                 & 98.1 $\pm$ 0.6      & \multicolumn{1}{c|}{$96.9 \pm 0.5$}   & 97.5 $\pm$ 0.5 & \multicolumn{1}{c|}{97.2 $\pm$ 0.5} & \multicolumn{1}{c|}{78.7 $\pm$ 3.6} & - \\
%RLSEP (U/S)                & $96.2 \pm 0.9$      & \multicolumn{1}{c|}{$95.7 \pm 1.4$}   & $95.9 \pm 0.7$ & \multicolumn{1}{c|}{$95.5 \pm 0.8$} & \multicolumn{1}{c|}{$66.6 \pm 5.6$} & - \\ \hdashline
CE (U)                   & $97.4 \pm 0.7$      & \multicolumn{1}{c|}{\textbf{98.0 $\pm$ 0.6}}   & $97.7 \pm 0.3$ & \multicolumn{1}{c|}{$97.4 \pm 0.4$} & \multicolumn{1}{c|}{$81.9 \pm 3.2$} & - \\
LSEP (U)                 & \textbf{98.6 $\pm$ 0.5}      & \multicolumn{1}{c|}{$97.3 \pm 0.5$}   & \textbf{97.9 $\pm$ 0.4} & \multicolumn{1}{c|}{\textbf{97.8 $\pm$ 0.4}} & \multicolumn{1}{c|}{\textbf{83.3 $\pm$ 3.5}} & - \\
RLSEP (U)                & $97.9 \pm 0.7$      & \multicolumn{1}{c|}{\textbf{98.0 $\pm$ 0.7}}   & \textbf{97.9 $\pm$ 0.6} & \multicolumn{1}{c|}{$97.7 \pm 0.6$} & \multicolumn{1}{c|}{$82.7 \pm 4.9$} & - \\ \hline
\end{tabular}
}
\caption{Statistical significance test results for each loss function, both in ranked (R) and unranked (U) evaluation metrics. For each loss function, we trained 30 networks with random initialization to get the means and standard deviations of the given metrics.}
\label{tab:significance}
\end{table}

\subsection{Implementation Details}

For all our experiments, we use MobileNetV3-Small \cite{DBLP:journals/corr/abs-1905-02244} as our scoring network, a 2-layer (576x128, 128xK) fully connected network for the threshold network, 64 batch size, SGD with learning rate 0.001 and a momentum of 0.9, and a weight decay of 1.e-5, for both networks. All of our trainings are done with maximum of 300 epochs and early stopping with patience 20.

\subsection{Metrics}
\textbf{Ranked Metrics.} For a ranked ground truth label set $R$ and label ranks $r_j$, we denote a pair set $P = \{(r_u, r_v)|(u,v) \in \{1, .., K\}^2, u < v, r_u \neq r_v \}$ which consists of each unique label pair for an input instance. Then we denote the pairs as positive if the left-hand rank is bigger than the right-hand rank, else they are denoted as negative. Due to the construction of the set, there are no equal-rank pairs, since there is no mechanism to evaluate if any two scores are equal or similar in our scope. For the score output $f(x_i; \theta)$, we construct a similar pair set using scores instead of ranks, and we again denote predicted positives and negatives in the same manner.\\
\textbf{Unranked Metrics.} For unranked evaluation, we have the ground truth positive label set $Y_i$. Predicted positive and negative labels are decided using the explained threshold method.

Precision, Recall, F1 and Accuracy scores for ranked, and unranked approaches are calculated using the positive, negative, predicted positive and predicted negative conventions described above, the scores are calculated for each input instance separately then averaged.\\
\textbf{0-1} denotes the exact match metric. For both ranked and unranked scenario, it yields the ratio of the number of instances that have no false positive or false negative labels to the number of all instances.\\
\textbf{mAP} denotes mean average precision metric. For a ground truth ranking label set $R_i$, iteratively the smallest rank in the ground truth set is changed to 0 until there is only one non-zero ranked label left, then for each iteration the ranked precision is calculated and averaged at the end.

\subsection{Comparison With Baselines}

\textbf{Ranked MNIST Dataset} We compare the performances of CE, LSEP, and RLSEP on ranked MNIST dataset by training a network with these loss functions. Particularly, we measure their performances for two problems: label ranking (represented with R) and multi-label classification (represented with U). Due to the nature of the loss function, RLSEP is trained with label ranking information, while LSEP and CE are trained without ranking information. To be able to calculate statistical significance tests, we repeat each experiment 30 times with different random network weight initializations, and report the mean and the standard error in Table \ref{tab:significance}. Due to limited computational resources, we use a subset of the ranked MNIST (5,000 training, 100 validation, and 100 test images) for this experiment. For the label ranking problem, our results demonstrate that RLSEP outperforms other methods in a statistically significant way in every metric ($p < 1.e-2$). For the multi-label classification problem, our method achieves either the second best or compatible results. This is in our opinion due to the extra ranking information the network has to consider in our optimization problem, which can also be regarded as an example to the "no free lunch" theorem. 

Additionally, we train the network model with CE, LSEP, and RLSEP with the full ranked MNIST dataset. The results are reported in Table \ref{tab:mnist_arc_results}. RLSEP achieves the best performances for the label ranking problem and achieves comparable performances for multi-label classification problem.

\textbf{ARC Dataset} Lastly, we experiment with the real world ARC dataset and compare our method, RLSEP, with LSEP and CE. We measure the label ranking (R) and multi-label classification (U) performances on the test split. As shown in the Table \ref{tab:mnist_arc_results}, RLSEP successfully utilizes ranking information and outperforms other methods in label ranking scenario, and it achieves comparable performance in multi-label classification scenario. Especially, it improves the performance of CE and LSEP in the hardest measure 0-1.

\begin{figure}[ht]
    \centering
    \hspace*{-0.09in}
    \includegraphics[width=0.35\textwidth]{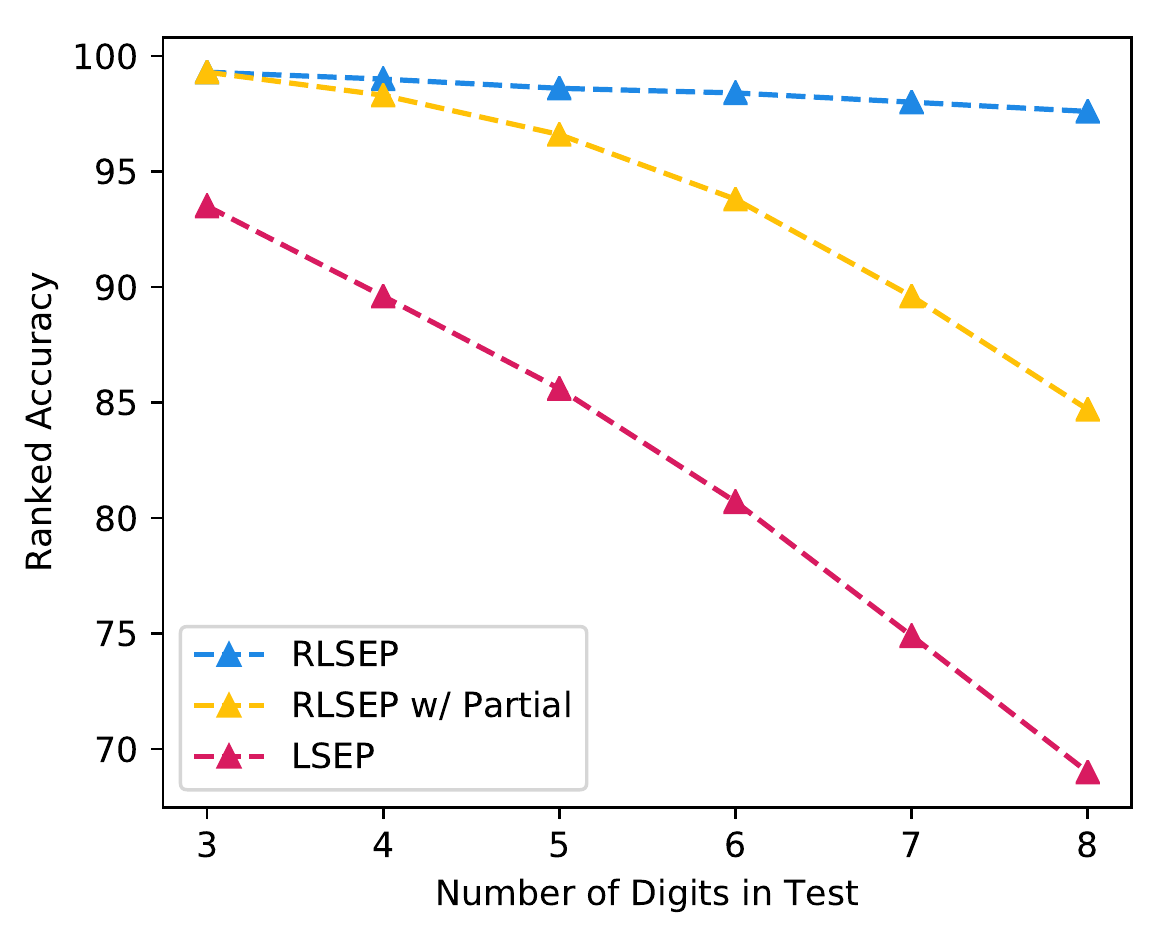}
    \caption{Illustration of ranked accuracy for three setups: RLSEP, RLSEP w/ Partial and LSEP. Each point in the x-axis is a different dataset with the corresponding number of varying size digits in each image. RLSEP and LSEP results are trained with the d-Digit dataset and tested with its test set, RLSEP w/ Partial is trained with 3-Digit dataset and tested on the d-Digit test sets.}
    \label{fig:increasing_labels}
\end{figure}
% \begin{figure}[ht]
%     \centering
%     \hspace*{-0.09in}
%     \includegraphics[width=0.45\textwidth]{big_fiery_rotaty.pdf}
%     \caption{Motivational but bigger}
%     \label{fig:big_fiery}
% \end{figure}

\begin{table*}[ht]
\renewcommand{\arraystretch}{1.2}
\centering
\begin{tabular}{c|cccccc|cccccc}
\multirow{2}{*}{Method} & \multicolumn{6}{c|}{Ranked MNIST} & \multicolumn{6}{c}{Architecture (ARC)} \\ \cline{2-13}
& Pr & \multicolumn{1}{c|}{Re} & F1 & \multicolumn{1}{c|}{Acc} & \multicolumn{1}{c|}{0-1} & mAP & Pr & \multicolumn{1}{c|}{Re} & F1 & \multicolumn{1}{c|}{Acc} & \multicolumn{1}{c|}{0-1} & mAP  \\ \cline{1-13}

CE (R) & \underline{$86.4$} & \multicolumn{1}{c|}{$87.1$} & $86.8$ & \multicolumn{1}{c|}{$87.6$} & \multicolumn{1}{c|}{\underline{$5.7$}}  & \underline{$80.4$} & \underline{$80.2$} & \multicolumn{1}{c|}{\underline{$81.8$}} & \underline{$81.0$} & \multicolumn{1}{c|}{\textbf{82.1}} & \multicolumn{1}{c|}{\underline{$12.8$}} & \underline{75.6}  \\

LSEP (R)                & $85.8$                             & \multicolumn{1}{c|}{\underline{$88.6$}} & \underline{$87.2$} & \multicolumn{1}{c|}{\underline{$87.8$}} & \multicolumn{1}{c|}{$5.6$}  & 79.3 & $79.8$ & \multicolumn{1}{c|}{$81.4$} & $80.6$ & \multicolumn{1}{c|}{$82.0$} & \multicolumn{1}{c|}{$9.6$} & \underline{75.6} \\

RLSEP (R)               & \textbf{98.5}      & \multicolumn{1}{c|}{\textbf{98.7}}   & \textbf{98.6}      & \multicolumn{1}{c|}{\textbf{98.8}}   & \multicolumn{1}{c|}{\textbf{72.6}}   & \textbf{95.9} & \textbf{80.4} & \multicolumn{1}{c|}{\textbf{82.3}} & \textbf{81.3} & \multicolumn{1}{c|}{\textbf{82.1}} & \multicolumn{1}{c|}{\textbf{14.0}} & \textbf{76.8}\\ \cline{1-13}

CE (U)                  & \textbf{99.1}      & \multicolumn{1}{c|}{\textbf{99.3}}   & \textbf{99.2}      & \multicolumn{1}{c|}{\textbf{99.1}}   & \multicolumn{1}{c|}{\textbf{94.3}}   & - & \textbf{58.2} & \multicolumn{1}{c|}{$73.3$} & $64.9$ & \multicolumn{1}{c|}{\textbf{79.7}} & \multicolumn{1}{c|}{\textbf{10.8}} & - \\

LSEP (U)                & \textbf{99.1}      & \multicolumn{1}{c|}{$98.7$} & $98.9$                              & \multicolumn{1}{c|}{\textbf{$98.8$}} & \multicolumn{1}{c|}{$91.8$} & - & $55.1$ & \multicolumn{1}{c|}{\textbf{81.4}} & \textbf{65.8} & \multicolumn{1}{c|}{\underline{$77.9$}} & \multicolumn{1}{c|}{$6.9$} & - \\

RLSEP (U)               & \textbf{99.1}      & \multicolumn{1}{c|}{\underline{$99.1$}} & {\underline{$99.1$}} & \multicolumn{1}{c|}{\underline{$99.0$}} & \multicolumn{1}{c|}{\underline{$93.3$}} & - & \underline{$55.3$} & \multicolumn{1}{c|}{\underline{$78.7$}} & \underline{$65.0$} & \multicolumn{1}{c|}{$77.8$} & \multicolumn{1}{c|}{\underline{$9.3$}} & - \\ \cline{1-13}

\end{tabular}
\caption{Performance results for both ranked (R) and unranked (U) scenarios and for the full Ranked MNIST and the ARC datasets. The best scores are depicted as bold, and the second-best scores are shown with an underline.}
\label{tab:mnist_arc_results}
\end{table*}

\subsection{Analysis of RLSEP}

%\textbf{Effects of number of labels} 
To further analyze RLSEP, we create different versions of the ranked MNIST where the number of labels per image vary. We report the label ranking accuracies of LSEP and RLSEP on these versions on Figure \ref{fig:increasing_labels}, with red and blue curves, respectively. Our results indicate that performance of LSEP drops significantly as the number of positive labels per image increases. On the other hand, the performance of RLSEP is robust to the number of positive labels. We conjecture that this is due to the lack of ranking information in LSEP, as the number of digits $T$ go up, the number of possible permutations $T!$ also goes up, decreasing the chance of random true positives dramatically.

Additionally, we experiment with partial training where the network model is trained with images that contains 3 digits only, and tested on images with different number of digits. While the label ranking accuracy of the network model trained with RLSEP loss drops as the number of digits increases as shown with the yellow curve in Figure \ref{fig:increasing_labels}, it still outperforms LSEP model that is trained with more digits. Even having access to partial rankings during training allows a model to reason with rankings better than a model that does not utilize ranking information at all.

% Figure \ref{fig:motivational} displays the change of scores calculated by RLSEP with respect to the changing scale factors of the digits in the image. Similarly in Figure \ref{fig:big_fiery}, we present the scores of the three loss functions we use in our work for the three images in Figure \ref{fig:motivational}, which are sampled at t=0.0, t=0.5 and t=1.0. It can be concluded that RLSEP successfully updates the scores as the digit sizes change and ranks the labels in a correct order, while LSEP does not respond to the changes in digit size as well as RLSEP and CE predicts almost equal scores for each digit regardless of the changes in size because of its nature.

\textbf{Calibration} To demonstrate the calibration performance, we create a new test set for the ranked MNIST dataset that consist of 10,000 images with 4 digits. Each image contains a single digit with scale 1, 2, 3, and 4. Examples can be seen in Figure \ref{fig:calibration} (bottom). We apply our model trained on full ranked MNIST dataset with RLSEP loss to the calibration test set and compare the scores of digits with different sizes. Figure \ref{fig:calibration} (top) shows that the model trained with RLSEP loss can output scores that are correlated with the scales of digits. We see that this correlation gets weaker as the scale of digits increases. We hypothesize that this is due to the increasing number of weights contributing to the score of the digit when it gets bigger, which increases the variation. Figure \ref{fig:calibration} (middle) depicts distinctive Gaussian distributions for different settings that further support our findings.

\subsection{Qualitative results}

Figure \ref{fig:mnist_results} shows our qualitative results on the Ranked MNIST. Examples a, b and c shows the accuracy of RLSEP compared to other loss functions. As can be seen, even if the number of digits are high or the scale factors are close, RLSEP still manages to give a correct prediction. On the other hand, examples d, e and f show challenging scenarios. In (e), the digits are not easily identifiable which affects the performance of all the networks, in (f) the digits are quite small and similar in size. It should be also noted that MNIST digits themselves vary in size, so the result is due to an expected kind of noise. Lastly (d) contains both hard to read and similar size digits. Figure~\ref{fig:arc_results} shows sample prediction results from the ARC dataset.

In brief, our qualitative results show that a network trained with RLSEP works excellently on regular examples and surpasses the other loss functions in terms of multi-label ranking. It also fails when there is either a complication which affects all the other loss functions (e.g. unrecognizable digits) or when it comes across an example in the range of expected error margin (e.g. digits outside the size difference which the dataset can represent).

\section{Conclusion}
In this paper, we proposed a novel loss function RLSEP based on pairwise ranking with respect to ordered ground truth labels, which is applied to multi-label ranking problem. We also introduced a new ranking dataset, the Ranked MNIST that is generated by changing sizes of MNIST digits. 

Our experiments provide strong evidence to that RLSEP is exceedingly successful on the task of multi-label ranking, and even with a dataset lacking the exact importance values of labels, it manages to learn well-calibrated scores. 

To our knowledge, our paper is the first contribution in multi-label ranking that includes label importance as ranks and incorporates that information into the training in terms of a new dedicated loss function. We demonstrated how this idea can be used in real world problems with an architectural facade dataset, in addition to the newly generated Ranked MNIST, which will be made available during publication.

\section{Acknowledgements}
We thank Prof. Aslı Cekmis Kanan for her support in collection and labeling of the ARC dataset. We thank Yusuf Hüseyin Şahin for his support in the making of this paper. We thank to Ata Gün Öğün for helping with some working environment tricks.

\bibliographystyle{named}
\bibliography{ijcai22}

\end{document}